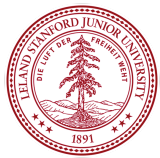

## Stanford University - CS230 Deep Learning

# Facial Expression Recognition with Deep Learning

*Improving on the State of the Art and Applying to the Real World*

Amil Khanzada
amilkh@stanford.edu

Charles Bai
cbai@stanford.edu

Ferhat Turker Celepcikay
turker@stanford.edu

*Abstract*—One of the most universal ways that people communicate is through facial expressions. In this paper, we take a deep dive, implementing multiple deep learning models for facial expression recognition (FER). Our goals are twofold: we aim not only to maximize accuracy, but also to apply our results to the real-world. By leveraging numerous techniques from recent research, we demonstrate a state-of-the-art 75.8% accuracy on the FER2013 test set, outperforming all existing publications. Additionally, we showcase a mobile web app which runs our FER models on-device in real time.

## I. INTRODUCTION

Facial expressions are fundamentally important in human communication. Although recognizing basic expressions under controlled conditions (e.g. frontal faces and posed expressions) is a solved problem with 98.9% accuracy, distinguishing basic expressions in natural conditions is still challenging due to variations in head pose, illumination, and occlusions [1-2].

However, with the advent of deep learning in the recent decade, FER technology under natural conditions has achieved remarkable accuracy in categorizing emotions from facial images, exceeding human level performance. This has allowed for the development of groundbreaking applications in sociable robotics, medical treatment, driver fatigue surveillance, and many other human-computer interaction systems [1].

In this paper, our goals were not only to better understand and improve the performance of emotion recognition models, but also to apply them to real world situations. We took several approaches from recent publications to improve accuracy, including transfer learning, data augmentation, class weighting, adding auxiliary data, and ensembling. In addition, we analyzed our models through error analysis and several interpretability techniques. We also applied our results to develop a mobile app to run our models on-device.

## II. RELATED WORKS

FER2013 was designed by Goodfellow et al. as a Kaggle competition to promote researchers to develop better FER systems. The top three teams all used CNNs trained discriminatively with image transformations [3]. The winner, Yichuan Tang, achieved a 71.2% accuracy by using the primal objective of an SVM as the loss function for training and additionally used the L2-SVM loss function [4]. This was a new development at the time and gave great results on the contest dataset.

There is a wealth of existing research in the FER domain. In particular, a recent survey paper on FER by S. Li and W. Deng sheds light on the current state of deep-learning-based approaches to FER [1]. Another paper by Pramerdorfer and Kampel [2] describes the approaches taken by six current state-of-the-art papers and ensembles their networks to achieve 75.2% test accuracy on FER2013, which is, to our knowledge, the highest reported in any published journal paper.

Among the six papers, Zhang et al. achieved the highest accuracy of 75.1% by employing auxiliary data and additional features: a vector of HoG features were computed from face patches and processed by the first FC layer of the CNN (early fusion). They also employed facial landmark registration, suggesting its benefits even in challenging conditions (facial landmark extraction is inaccurate for about 15% of images in the FER dataset) [5]. The paper with the second highest accuracy by Kim et al. utilized face registration, data augmentation, additional features, and ensembling [6].

From our graduate community at Stanford, we also found reports from recent CS229 and CS230 projects on FER useful as reference [7,8].

## III. DATASETS

FER is a well-studied field with numerous available datasets. We used FER2013 as our main dataset and drove up accuracy on its test set by using CK+ and JAFFE as auxiliary datasets. We also created our own web app dataset to tune our models to work better in real world scenarios.

**FER2013 Dataset**

FER2013 is a well-studied dataset and has been used in ICML competitions and several research papers. It is one of the more challenging datasets with human-level accuracy only at 65±5% and the highest performing published works achieving 75.2% test accuracy. Easily downloadable on Kaggle, the dataset's 35,887 contained images are normalized to 48x48 pixels in grayscale. FER2013 is, however, not a balanced dataset, as it contains images of 7 facial expressions, with distributions of Angry (4,953), Disgust (547), Fear (5,121), Happy (8,989), Sad (6,077), Surprise (4,002), and Neutral (6,198) [3].

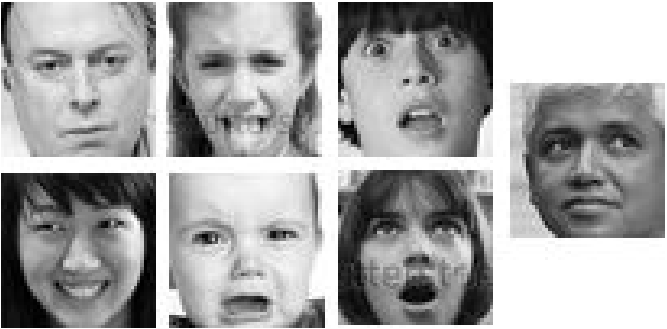

Figure 1: Images from each emotion class in the FER2013 dataset.

**Japanese Female Facial Expression Dataset**

The Japanese Female Facial Expression (JAFFE) is a relatively small dataset containing 213 images of 10 Japanese female models. The images are labeled with 7 facial expressions, as in FER2013.

**Mobile Web App Dataset**

To properly train our web app model, we found it necessary to gather images taken from the app itself. We employed several methods including Facebook posts, video tutorials, and face-to-face requests, eventually gathering 258 labeled images from 12 people. Although our dataset was ethnically imbalanced with respect to FER2013 (the majority of our training data), it was sufficient to meet the satisficing accuracy metrics of our web app after we had employed the appropriate dev/test distributions.

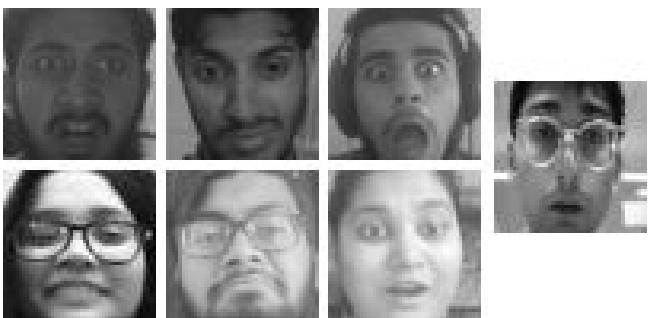

Figure 2: Images of each emotion class in our web app dataset.

**Extended Cohn-Kanade Dataset**

The extended Cohn-Kanade dataset (CK+) contains 593 image sequences of 123 subjects between the ages of 18 to 50. Each sequence of images contains 10 to 60 frames of a subject transitioning from neutral to the target emotion and each frame is roughly 640x480 with grayscale and/or color values [9].

## IV. MODELS

**Baseline Model**

In order to better understand the problem, we decided to first try to tackle this problem from scratch, building a vanilla CNN using four 3x3x32 same-padding, ReLU filters, interleaved with two 2x2 MaxPool layers, and completed with a FC layer and softmax layer. We also added batchnorm and 50% dropout layers to address high variance and improve our accuracy from 53.0% to 64.0%.

**Five-Layer Model**

One of the highest accuracy papers we could find was Pramerdorfer and Kampel's [2], which reported 75.2% accuracy, despite not using auxiliary training data or facial landmark registration. The authors achieved these results by studying six other papers and ensembling their networks. Because of the simplicity of the network, we decided to replicate their exercise of reproducing the results of Kim et al. [6].

This model consists of three stages of convolutional and max-pooling layers, followed by an FC layer of size 1024 and a softmax output layer. The convolutional layers use 32, 32, and 64 filters of size 5x5, 4x4, and 5x5, respectively. The max-pooling layers use kernels of size 3x3 and stride 2. ReLU was utilized as the activation function. To improve performance, we also added batchnorm at every layer and 30% dropout after the last FC layer. To fine tune the model, we trained it for 300 epochs, optimizing the cross-entropy loss using stochastic gradient descent with a momentum of 0.9. The initial learning rate, batch size, and weight decay are fixed at 0.1, 128, and 0.0001, respectively. The learning rate is halved if the validation accuracy does not improve for 10 epochs.

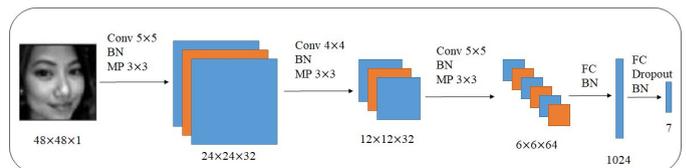

Figure 3: Architecture of Kim et al.'s five-layer model [6].

**Transfer Learning**

Since the FER2013 dataset is quite small and unbalanced, we found that utilizing transfer learning significantly boosted the accuracy of our model. We explored transfer learning, using the Keras VGG-Face library and each of ResNet50, SeNet50 and VGG16 as our pre-trained models.

To match the input requirements of these new networks which expected RGB images of no smaller than 197x197, we resized and recolored the 48x48 grayscale images in FER2013 during training time.

**Fine Tuning ResNet50**

ResNet50 is the first pre-trained model we explored. ResNet50 is a deep residual network with 50 layers. It is defined in Keras with 175 layers. We started off by reproducing work from Brechet et al. [10]. We replaced the original output layer with two FC layers of sizes 4,096 and 1,024 respectively and a softmax output layer of 7 emotion classes. We froze the first 170 layers in ResNet, and kept the rest of the network trainable. We used SGD as our optimizer with a learning rate of 0.01 and a batch size of 32. After training for 122 epochs using SGD with a 0.01 learning rate and a batch size of 128, we achieved 73.2% accuracy on the test set.

We also tried to freeze the entire pre-trained network and only train the FC layers and output layer, but the model failed to fit onto the training set in the first 20 epochs despite our many attempts to adjust hyperparameters. Given our limited computational resources, we decided to not explore further on this route.

**Fine Tuning SeNet50**

SeNet50 is another pre-trained model we explored. It is a deep residual network with 50 layers. SeNet50 has a similar structure with ResNet50, so we didn't spend too much time tuning this model. We trained the network on the set of parameters we used for ResNet50 and achieved 72.5% accuracy on the test set.

**Fine Tuning VGG16**

Although much shallower than ResNet50 and SeNet50 with only 16 layers, VGG16 is more complex and has many more parameters. We kept all pre-trained layers frozen and added two FC layers of size 4096 and 1024 respectively with 50% dropout. After 100 epochs of training with the Adam optimizer, we achieved an accuracy of 70.2% on the test set.

## V. MOBILE WEB APP

Rather than take a purely theoretical approach, we thought it would be challenging and novel to apply our work to the real world by developing a mobile web app to run our model on-device in real-time.

Given the memory, disk, and computational limitations of mobile devices, we carefully considered the appropriate evaluation metrics for our model and agreed that low memory/disk requirements and fast prediction speeds were far more important than small gains in accuracy. We chose a satisficing metric of 100ms recognition speed on-device and kept accuracy as our optimizing metric. This prompted us to research shorter networks and eventually adopt B.-K. Kim et al.'s five-layer CNN model [6].

Tuning the model to perform well on our app presented several challenges - particularly dataset mismatch. Unlike those in our training datasets, images captured by the web app often had poor illumination and tilted angles. We overcame this by randomly shuffling 80% of our web app dataset into the training set along with all images from the other datasets and keeping 20% of our web app dataset in the test set. After training for 120 epochs without modification to the hyperparameters, we achieved an accuracy of 69.8% on the web app test set with 40ms recognition speed, which was sufficient for our evaluation metrics.

Architecturally, our web app is hosted on Firebase and utilizes TensorFlow.js, React.js, and face-api.js to detect, crop, and resize the user's face, before feeding it as a 48x48 image with one grayscale channel to our model. Additionally, to reduce disk space and memory footprint, model weights are shrunk using tensorflowjs_converter before download to the user's device.

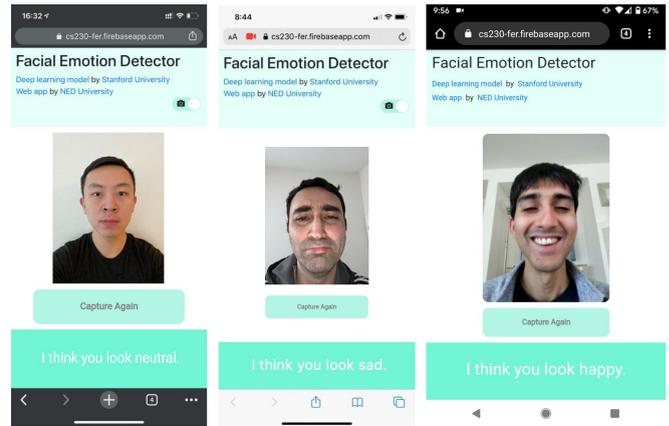

Figure 4. Screenshots of the mobile web app making predictions.

## VI. METHODS

**Auxiliary Data & Data Preparation**

Although several FER datasets are available online, they vary widely in image size, color, and format, as well as labeling and directory structure. We addressed these differences by simply partitioning all input datasets into 7 directories (one for each class). During training, we loaded images in batches from disk (to avoid memory overflow) and utilized Keras data generators to automatically resize and format the images.

**Data Augmentation**

We researched and experimented with commonly used techniques in existing FER papers and achieved our best results with horizontal mirroring, ±10 degree rotations, ±10% image zooms, and ±10% horizontal/vertical shifting.

**Class Weighting**

To alleviate the class imbalance problem, we applied class weighting inversely proportional to the number of samples. For the disgust class, we were able to drop the misclassification rate from 61% to 34%.

**SMOTE**

The Synthetic Minority Over-sampling Technique (SMOTE) involves oversampling minority classes and undersampling majority classes to get the best results [11]. Although utilizing SMOTE resulted in a perfectly balanced training dataset, our models quickly overfit to the training dataset and we decided not to experiment further.

**Ensembling & Test-Time Augmentation (TTA)**

We performed ensembling with soft voting of seven models to significantly improve our highest test accuracy from 73.2% to 75.8%. Similarly, TTA with horizontal flip and seven augmented images improved the test accuracy of our five-layer model by 1.7%.

## VII. RESULTS / DISCUSSION

**Accuracy-Driven Models**

Table 1 shows the accuracies our best models achieved on the FER2013 private test dataset. Results from Tang [4] (the Kaggle competition winner) and Pramerdorfer et al. [2] (the highest published accuracy) are also depicted.

Table 1: Network parameters and results on FER2013 private test set.

| Model | Depth | Parameters | Accuracy |
|---|---|---|---|
| (Human-level) | - | - | 65±5% |
| Tang [4] | 4 | 12(m) | 71.2% |
| Pramerdorfer et al. [2] | 10/16/33 | 1.8/1.2/5.3(m) | 75.2% |
| Baseline | 5 | 37.8(m) | 64.0% |
| Five-Layer Model | 5 | 2.4(m) | 66.3% |
| VGG16 | 16 | 138(m) | 70.2% |
| SeNet50 | 50 | 27(m) | 72.5% |
| ResNet50 | 50 | 25(m) | 73.2% |
| **Ensemble** | - | - | **75.8%** |

Most of the publications which achieved state-of-the-art accuracies on FER2013 utilized auxiliary training data [12-13]. Table 2 demonstrates our accuracy gains from employing auxiliary data with care taken to avoid dataset bias. It also depicts our success in implementing class weighting, which significantly increased accuracies on frequently misclassified emotions. Ensembling several models with and without class weights improved the overall accuracy.

Table 2: Accuracies for different methods with and without auxiliary data.
*NCW* = no class weights / *WCW* = with class weights

| Dataset | ResNet50 | | SeNET50 | | VGG16 | | Ensemble |
|---|---|---|---|---|---|---|---|
| | NCW | WCW | NCW | WCW | NCW | WCW | |
| FER2013 | 73.2% | 67.7% | 70.0% | 68.9% | 69.5% | 70.0% | 74.8% |
| Auxiliary | 72.7% | 72.4% | 72.5% | 71.6% | 70.2% | 69.6% | 75.8% |

Given the high complexity of transfer learning models and relatively small size of our datasets, we also experienced overfitting while training. Shown in Figure 5, although we added 50% dropout for the last three layers, our ResNet50 transfer learning model quickly overfit to the training set with dev accuracy starting to flatten after only 30 epochs.

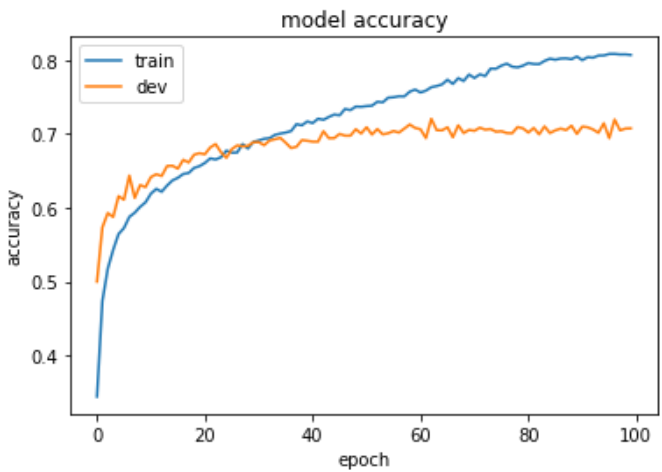

Figure 5: Learning curve for ResNet50 WCW transfer learning model.

**Web App Model**

For our five-layer web app model, we achieved a test set accuracy of 69.8% and recognition speed of 40ms.

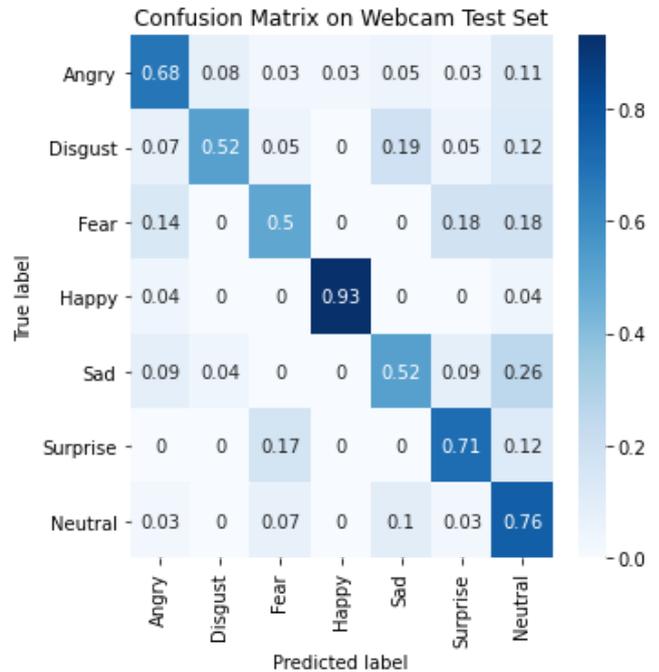

Figure 6: Confusion matrix of the web app model.

For error analysis, we targeted confusion matrix cells with high misclassifications. One interesting example was an image labeled fear that was classified by our model as angry (29%), fear (28%), and sad (26%), similar to mispredictions by humans on the same image.

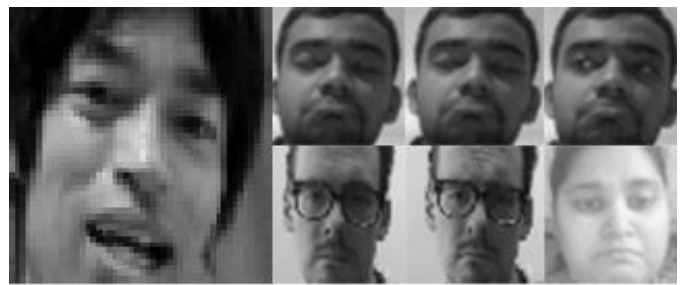

Figure 7: One fear and six sad images misclassified by the web app model.

We also investigated the high misclassification of sad images as neutral, where one of the subjects was misclassified on all of his sad images. Recognizing this as a class imbalance problem, we overcame it by collecting more sad images for our web app dataset.

It is also worth mentioning that through the error analysis process, we identified and manually corrected three images that were mislabeled in our web app data set.

**Interpretability**

To better understand our network's behavior, we employed various methods including Grad-CAM [14], XRAI [15], vanilla gradients, and occlusion maps [16].

Running occlusion maps on correctly classified images in our web app model, we observed the network had learned to focus on the nose and mouth to make predictions for disgust, the mouth for happiness, and the eyes and nose for surprise. For neutral images, we found that the network focused on all parts of the face except for the nose, which made sense given that small changes in non-nose regions tend to correspond to emotion changes.

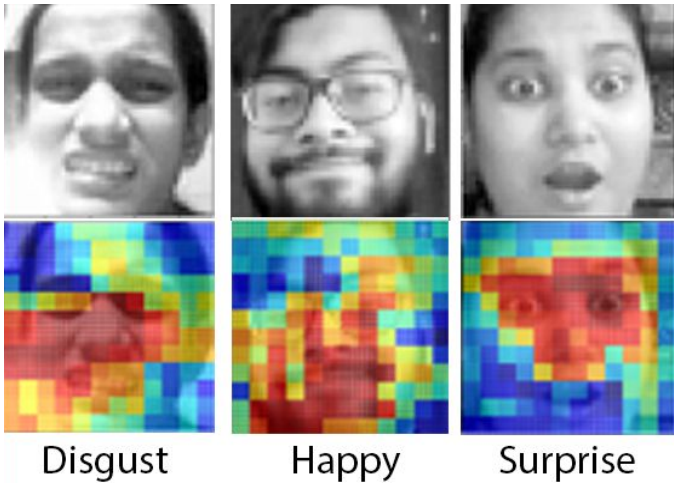

Figure 8: Occlusion-based heatmaps for disgust, happy, and surprise. Pixels assigned warmer colors have greater importance in the prediction.

**Challenges**

It is worth mentioning that because our models exceeded human-level accuracy, error analysis was particularly challenging for some misclassifications, such as the fear image discussed prior.

Additionally, because emotions are highly subjective, Bayes error is high and it is often the case that an image can have multiple interpretations as shown in Figure 9 [17].

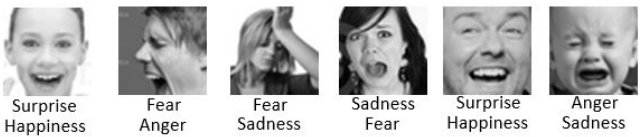

Figure 9: FER2013 images with two possible labels.

## VIII. CONCLUSION

When we started this project, we had two goals, namely, to achieve the highest accuracy and to apply FER models to the real world. We explored several models including shallow CNNs and pre-trained networks based on SeNet50, ResNet50, and VGG16. To alleviate FER2013's inherent class imbalance, we employed class weights, data augmentation, and auxiliary datasets. By ensembling seven models we achieved 75.8% accuracy, which is the highest to our knowledge. We also found through network interpretability that our models learned to focus on relevant facial features for emotion detection.

Additionally, we demonstrated that FER models could be applied in the real world by developing a mobile web application with real-time recognition speeds. We overcame data mismatch issues by building our own training dataset and also tuned our architecture to run on-device with minimal memory, disk, and computational requirements.

## IX. FUTURE WORK

To further improve the accuracy of our models, we hope to utilize facial landmark detection and alignment [6], implement attentional CNNs [16], and retrain our network by occluding facial features irrelevant to emotion recognition [16]. We would also like to employ more auxiliary data (in particular AffectNet which contains over a million labeled images) and balance our training dataset with methods such as ADASYN [18]. Furthermore, we believe there is great potential for improvement with pipeline models, where commonly misclassified emotion pairs (e.g. neutral and sad) are fed to secondary networks with higher accuracy rates between those specific emotions.

To further adapt our models to the real world, we hope to integrate contemporary Psychological research, particularly the arousal-valence emotional model [19], and also utilize multi-label classification to better handle images with multiple possible emotion labels [17]. We would like to improve the robustness and accuracy of our web app model by increasing the web app dataset's size and also applying different data augmentation techniques to address varying camera brightness and angle issues. Additionally, we would like to apply our work to benefit humanity, such as by employing it to support shared empathy. We hope to submit our results to conferences such as NeurIPS and take part in competitions similar to FER2013.

Finally, we have started the Pakistani Female Facial Expression dataset project (PKFFE.org) in hopes to address the heavy ethnicity bias issues of existing facial datasets.

## X. CODE

We have open sourced our work, including the mobile web app, under the MIT license for the benefit of academia. For confidentiality reasons, we have kept our web app training dataset private. A poster and video showcasing our work can also be found in our GitHub repository.

https://github.com/amilkh/cs230-fer

## XI. MOBILE WEB APP

Our mobile web application can be accessed below.

http://cs230-fer.firebaseapp.com/

## XII. CONTRIBUTIONS

**Amil Khanzada**
- Managing project and external contributors
- Mobile web app development and model tuning
- Auxiliary data and dataset preprocessing
- Error analysis and network interpretability

**Charles Bai**
- Transfer learning models
- Hyperparameter tuning
- Dataset preprocessing
- Infrastructure management

**Turker**
- Transfer learning models
- Researching and tuning shallow models
- Dataset preprocessing
- Data augmentation
- Class weighting and SMOTE
- Ensembling and TTA

## XIII. ACKNOWLEDGEMENTS

We are very grateful to Assistant Professor Asma Khan and her team of undergraduate students in the Software Engineering department of NED University in Karachi, Pakistan for helping us to apply our work to the real world. The mobile web application development was led by Muhammad Bilal Khan (app/hosting), Muhammad Hassan-ur-Rehman (app/hosting), and Tooba Ali (UI design). Muhammad Ashhad Bin Kashif and Summaiya Sarfaraz assisted us with network interpretability. Muhammad Hasham Khalid and Midha Tahir spent several hours pair programming with us on auxiliary dataset preparation, error analysis, and model tuning for the web app.

We appreciate Pawan Nandakishore, data scientist at Colaberry, for his regular guidance over the course of the project, particularly in research of related works, interpretability, and model tuning.

Finally, we owe our success to our Stanford CS230 teaching assistant Chris Waites for providing us weekly mentoring, support, and feedback on our project. And, last but not least, we are indebted to our Professors Andrew Ng and Kian Katanforoosh.